\documentclass[accepted]{uai2023} %

\usepackage[american]{babel}

\usepackage{natbib} %
\usepackage{mathtools} %
\usepackage{amsfonts}
\usepackage{booktabs} %
\usepackage{tikz} %

\usepackage{pgf, pgfplots}
\pgfplotsset{compat=1.18}
\usetikzlibrary{shapes,arrows.meta, calc, positioning,shapes}
\usepackage{acronym}
\usepackage{bm}
\usepackage{mleftright}
\usepackage{makecell}

\newcommand\vek[1]{{\bm{#1}}}

\DeclareMathOperator*{\argmin2}{arg\,min}

\definecolor{KITgreen}{rgb}{0,.59,.51}
\definecolor{KITpalegreen}{RGB}{130,190,60} 
\definecolor{KITblack}{rgb}{0,0,0}
\definecolor{KITblue}{rgb}{.27,.39,.66}
\definecolor{KITred}{rgb}{.63,.13,.13}
\definecolor{KITpurple}{rgb}{.64,.06,.48}
\definecolor{KITcyan}{rgb}{.14,.63,.87}
\definecolor{KITyellow}{rgb}{.98,.89,0}
\definecolor{KITorange}{rgb}{.87,.60,.10}

\acrodef{AWGN}[AWGN]{additive white Gaussian noise}
\acrodef{BMI}[BMI]{bitwise mutual information}
\acrodef{BP}[BP]{belief propagation}
\acrodef{BPSK}[BPSK]{binary phase-shift keying}
\acrodef{CCCP}[CCCP]{concave-convex procedure}
\acrodef{GNN}[GNN]{graph neural network}
\acrodef{iid}[i.i.d.]{independent and identically distributed}
\acrodef{ISI}[ISI]{inter-symbol interference}
\acrodef{KL}[KL]{Kullback-Leibler}
\acrodef{LLR}[LLR]{log-likelihood ratio}
\acrodef{NN}[NN]{neural network}
\acrodef{SPA}[SPA]{sum-product algorithm}

\title{Local Message Passing on Frustrated Systems}

\author[1]{\href{mailto:<luca.schmid@kit.edu>?Subject=Your UAI 2023 paper}{Luca Schmid}{}}
\author[1]{Joshua Brenk}
\author[1]{Laurent Schmalen}
\affil[1]{%
    Communications Engineering Lab (CEL)\\
    Karlsruhe Institute of Technology (KIT)\\
    Karlsruhe, Germany
}
  
\begin{document}
\maketitle

\begin{abstract}
    Message passing on factor graphs is a powerful framework for probabilistic inference, which finds important applications in various scientific domains.
    The most wide-spread message passing scheme is the sum-product algorithm (SPA) which gives exact results on trees but often fails on graphs with many small cycles. We search for an alternative message passing algorithm that works particularly well on such cyclic graphs. Therefore, we challenge the extrinsic principle of the SPA, which loses its objective on graphs with cycles. We further replace the local SPA message update rule at the factor nodes of the underlying graph with a generic mapping, which is optimized in a data-driven fashion. These modifications lead to a considerable improvement in performance while preserving the simplicity of the SPA. We evaluate our method for two classes of cyclic graphs: the $2 \times 2$ fully connected Ising grid and factor graphs for symbol detection on linear communication channels with inter-symbol interference. 
    To enable the method for large graphs as they occur in practical applications, we develop a novel loss function that is inspired by the Bethe approximation from statistical physics and allows for training in an unsupervised fashion.
\end{abstract}

\section{Introduction}\label{sec:intro}
Message passing on graphical models is a powerful framework to efficiently solve inference and optimization problems. The most prominent message passing algorithm is the \ac{SPA}, also known as \ac{BP}~\citep{pearl_probabilistic_1988}, which implements exact inference on tree-structured graphs~\citep{kschischang_factor_2001}. Due to its simplicity, the \ac{SPA} is often applied to cyclic graphs where it becomes an iterative and approximate algorithm. While this works surprisingly well for various applications, such as decoding of low-density parity-check codes~\citep{gallager_ldpc_1963}, a class of error-correcting codes, the \ac{SPA} performs poorly on frustrated systems, i.e., on graphs with many cycles and strong coupling between the nodes.

The seminal work of \citet{yedidia_generalized_2000} revealed a connection between the \ac{SPA} and free energy approximations of statistical physics, in particular, the fixed points of \ac{BP} correspond to stationary points of the Bethe free energy.
Based on this insight, alternative message passing methods were proposed which directly minimize the Bethe free energy \citep{yuille_cccp_2002, welling_belief_2013}. These algorithms are guaranteed to converge to an extremum of the Bethe free energy but are computationally more demanding than plain \ac{BP}.
\citet{wainwright_tree-reweighted_2003} proposed tree-reweighted \ac{BP} as a message passing algorithm on the ``convexified'' Bethe free energy, which is guaranteed to have a global minimum. While this algorithm has stronger convergence guarantees compared to \ac{BP}, it involves the selection and optimization of so-called edge appearance probabilities, a graph-specific problem that is often non-trivial for practical applications.
\citet{yedidia_generalized_2000} proposed ``generalized \ac{BP}'' as an algorithm that passes messages between regions of nodes instead of single nodes. Larger regions will generally improve the quality of the approximation, however, they also increase the computational complexity.

Recently, model-based deep learning has shown great potential to empower various suboptimal algorithms, such as the \ac{SPA} on cyclic graphs. Neural \ac{BP}, proposed by~\citet{nachmani_learning_2016}, unfolds the iterations of the \ac{SPA} on its underlying graph and equips the resulting deep network  with tunable weights. The GAP algorithm of~\citet{schmid_low-complexity_2022} varies the observation model by preprocessing, thereby shaping a graph with more favorable properties  with respect to \ac{BP} performance. 
\citet{satorras_neural_2021} extend \acp{GNN} to factor graphs and propose a hybrid model where \ac{BP} runs conjointly to a \ac{GNN} which is structurally identical to the original factor graph but has fully parametrized message updates. 
All these works have in common that they are based on the \ac{SPA} as a core concept which is vigorously improved using machine learning in order to compensate for its shortcomings on graphs with cycles.
In this work, we follow an alternative approach and directly search for alternative message passing algorithms that perform especially well on graphs with cycles, where the \ac{SPA} tends to fail. To this end, we replace the well-known \ac{SPA} message update rule with a compact \ac{NN}, which is optimized to find a superior local message update rule.
Furthermore, we discuss the role of the extrinsic information principle which was originally introduced for tree-structured graphs.
Based on the close connection of \ac{BP} to the Bethe approximation, we propose a novel end-to-end loss function that allows unsupervised and application-agnostic training of new message passing schemes.

\section{Background}
We briefly introduce factor graphs and the \ac{SPA} as a widespread framework for probabilistic inference on graphical models. We refer the reader to~\citep{kschischang_factor_2001} for an excellent in-depth treatment of the topic.
\subsection{Factor Graphs}
Let $f(\mathcal{X})$ be a multivariate function of ${\mathcal{X}=\{x_1,\ldots,x_N \}}$ which factors into a product of \emph{local} functions $f_j$:
\begin{equation}
    f(\mathcal{X}) = \frac1Z \prod\limits_{j=1}^J f_j(\mathcal{X}_j),\quad \mathcal{X}_j \subseteq \mathcal{X}.
    \label{eq:factorization}
\end{equation}

A \emph{factor graph} visualizes the factorization in \eqref{eq:factorization} as a bipartite graph. 
Every variable $x_n$ is represented by a unique vertex, a so-called variable node, which we draw as a circle in the graph. Factor nodes represent the local functions~$f_j$ and are visualized by squares. The undirected edges of the graph connect a factor node~$f_j(\mathcal{X}_j)$ with a variable node $x_n$ if and only if $f_j$ is a function of~$x_n$, i.e., if $x_n \in \mathcal{X}_j$. From a graphical perspective, $\mathcal{X}_j$ thus corresponds to the set of adjacent variable nodes to the factor node $f_j$. Similarly, we define $\mathcal{N}(x_n)$ to be the set of adjacent factor nodes to the variable node $x_n$. 

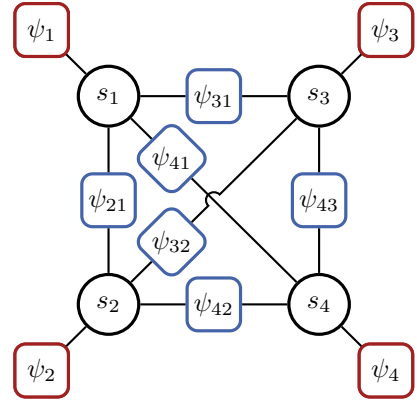
\begin{figure}[tb]
\centering
    \tikzstyle{fn} = [draw, very thick, rectangle, inner sep=1em, rounded corners]
    \tikzstyle{vn} = [draw, very thick, circle, inner sep=.8em]
    \tikzset{jump/.style args={(#1) to (#2) over (#3) by #4}{
        insert path={
            let \p1=($(#1)-(#3)$), \n1={veclen(\x1,\y1)}, 
            \n2={atan2(\y1,\x1)}, \n3={abs(#4)}, \n4={#4>0 ?180:-180}  in 
            (#1) -- ($(#1)!\n1-\n3!(#3)$) 
            arc (\n2:\n2+\n4:\n3) -- (#2)
        }
     }}
    \begin{tikzpicture}[auto, node distance=5.5em and 5.5em, thick]
    	\node[vn,label=center:$s_1$](c1){};
    	\node[vn, right=of c1,label=center:$s_3$](c3){};
    	\node[vn, below=of c1,label=center:$s_2$](c2){};
    	\node[vn, right=of c2,label=center:$s_4$](c4){};
    	\node[fn, color=KITblue, anchor=center, label=center:$\psi_{31}$](I31) at ($(c1)!0.5!(c3)$) {};
    	\node[fn, color=KITblue, anchor=center, label=center:$\psi_{21}$](I21) at ($(c1)!0.5!(c2)$) {};
    	\node[fn, color=KITblue, anchor=center, label=center:$\psi_{43}$](I43) at ($(c3)!0.5!(c4)$) {};
    	\node[fn, color=KITblue, anchor=center, label=center:$\psi_{42}$](I42) at ($(c2)!0.5!(c4)$) {};
    	\node[fn, color=KITblue, anchor=center, label=center:$\psi_{32}$, rotate=45](I32) at ($(c2)!0.3!(c3)$) {};
        \node[fn, color=KITblue, anchor=center, label=center:$\psi_{41}$, rotate=45](I41) at ($(c1)!0.3!(c4)$) {};
    
    	\node[fn, color=KITred, above left=0.6em and 0.6em of c1, label=center:$\psi_{1}$](F1){};
    	\node[fn, color=KITred, below left=0.6em and 0.6em of c2, label=center:$\psi_{2}$](F2){};
    	\node[fn, color=KITred, above right=0.6em and 0.6em of c3, label=center:$\psi_{3}$](F3){};
    	\node[fn, color=KITred, below right=0.6em and 0.6em of c4, label=center:$\psi_{4}$](F4){};
    
    	\draw[-] (c1) -- (F1.center);
    	\draw[-] (c2) -- (F2.center);
    	\draw[-] (c3) -- (F3.center);
    	\draw[-] (c4) -- (F4.center);
    	\draw[-,black] (c1) -- (I31);
    	\draw[-,black] (c3) -- (I31);
    	\draw[-,black] (c2) -- (I42);
    	\draw[-,black] (c4) -- (I42);
    	\draw[-,black] (c1) -- (I21);
    	\draw[-,black] (c2) -- (I21);
        \draw [jump=(c3) to (I32) over ($(c2)!0.5!(c3)$) by 3pt];
    	\draw[-,black] (c2) -- (I32);
    	\draw[-,black] (c3) -- (I43);
    	\draw[-,black] (c4) -- (I43);
        \draw[-,black] (c1) -- (I41);
    	\draw[-,black] (c4) -- (I41);

        \node[fn, color=KITred, fill=white, above left=0.6em and 0.6em of c1, label=center:$\psi_{1}$](F1){};
    	\node[fn, color=KITred, fill=white, below left=0.6em and 0.6em of c2, label=center:$\psi_{2}$](F2){};
    	\node[fn, color=KITred, fill=white, above right=0.6em and 0.6em of c3, label=center:$\psi_{3}$](F3){};
    	\node[fn, color=KITred, fill=white, below right=0.6em and 0.6em of c4, label=center:$\psi_{4}$](F4){};
    \end{tikzpicture}
    \caption{Factor graph representation of \eqref{eq:mrf} with factor nodes of degree~2 (blue) and degree~1 (red). This graph also models the $2 \times 2$ fully connected Ising graph of Sec.~\ref{sec:examples}.}
    \label{fig:Ising2x2}
\end{figure}

In this work, we restrict the variables ${x_n \in \{+1,-1\}}$ to be binary and the local factors $f_j$ to be either functions of a singular variable~$x_n$ or functions of pairs~$(x_n,x_m)$ such that the factorization becomes
\begin{equation}
    f(\mathcal{X}) = \frac1Z \prod\limits_{n=1}^N \psi_n(x_n) \prod_{\mathclap{(n,m)\in \mathcal{E}}} \psi_{n,m}(x_n,x_m),
    \label{eq:mrf}
\end{equation}
where $\mathcal{E}$ is the set of edges in the graph. Figure~\ref{fig:Ising2x2} shows an exemplary factor graph.

\subsection{Sum-product Algorithm}
The \ac{SPA} is a message passing algorithm that operates in a factor graph and attempts to determine the marginals of the multivariate function~$f(\mathcal{X})$.
Messages are propagated between the nodes of the factor graph along its edges and represent interim results of the marginalization.
Let $m_{f_j \rightarrow x_n}(x_n)$ denote a message sent from a factor node~$f_j$ along an edge to a variable node~$x_n$ and let $m_{x_n \rightarrow f_j}(x_n)$ denote a message on the same edge, but sent in the opposite direction.
If the factor graph visualizes a probabilistic model, i.e., if the variable nodes represent random variables, a message $m_{f_j \rightarrow x_n}(x_n)$ can be interpreted as a probabilistic statement from node $f_j$ about the random variable $x_n$ to be in one of its possible states \citep{yedidia_constructing_2005}.
The \ac{SPA} defines the updates of the propagating messages at the nodes of the factor graph according to the simple rules~\citep{kschischang_factor_2001}:
\begin{align}
    m_{x \rightarrow f_j}(x) &= \prod_{f_i \in \mathcal{N}(x) \setminus f_j} m_{f_i \rightarrow x}(x) \label{eq:VN_update_generic} \\
    m_{f_j \rightarrow x}(x) &= \sum_{\sim \{ x \}} \left( f_j(\mathcal{X}_j) \prod_{x' \in \mathcal{X}_j \setminus x} m_{x' \rightarrow f_j}(x')  \right). \label{eq:FN_update_generic}
\end{align}
The summary operator $\sum_{\sim \{x\}}$ denotes the marginalization over all variables in $\mathcal{X}_j$ except for $x$.
One key property of the \ac{SPA} is the \emph{extrinsic information principle} which states that the update of an outgoing message $m_{\text{A} \rightarrow \text{B}}$ at node $\text{A}$ destined to node $\text{B}$ does not depend on the incident message $m_{\text{B} \rightarrow \text{A}}$ which travels on the same edge but in opposite direction. 
For the special case of degree-2 factor nodes $\psi_{n,m}(x_n,x_m)$, the \ac{SPA} update rule~\eqref{eq:FN_update_generic} thereby simplifies to
\begin{equation*}
m_{\psi_{n,m} \rightarrow x_n}(x_n) = \sum_{x_m}\psi_{n,m}(x_n,x_m) \cdot m_{x_m \rightarrow \psi_{n,m}}(x_m).
\end{equation*}
Messages at factors nodes $\psi_n(x_n)$ with degree~1 are not updated at all.

Initially, all messages are set to some unbiased state before they are iteratively updated according to a certain schedule. 
For tree-structured graphs, the messages converge after they have once traveled forward and backward through the entire graph. The result of the \ac{SPA}, i.e., the marginal functions~$f(x_n)$, are finally obtained by a combination of all messages incident to the respective variable nodes:
\begin{equation*}
    f(x_n) = \prod_{f_i \in \mathcal{N}(x_n)} m_{f_i \rightarrow x_n}(x_n).
\end{equation*}
Since the \ac{SPA} makes no reference to the topology of the factor graph and the message updates are local, the \ac{SPA} may also be applied to factor graphs with cycles~\citep{yedidia_understanding_2003}.
On graphs with cycles, the \ac{SPA} only yields an approximation of the exact marginals. While this approximation works surprisingly well in many cases, even including particular classes of graphs with many small cycles, there are also cases where the results are quite poor or where the \ac{SPA} does not converge at all~\citep{murphy_loopy_1999}.

\paragraph{Relation to the Bethe Approximation}
In their seminal work, \citet{yedidia_generalized_2000} showed a revealing connection between the \ac{SPA} and free energy approximations in statistical physics.
From a variational perspective, probabilistic inference can be seen as an optimization problem
\begin{equation}
    q = \arg\min_{q \in \mathbb{M}} D_\text{KL}(q || p),   \label{eq:prob_inference}
\end{equation}
where we want to find the distribution $q$ from the set~$\mathbb{M}$ of all globally valid probability distributions, known as the \emph{marginal polytope}~\citep{wainwright_graphical_2008}. Since the \ac{KL} divergence $D_\text{KL}(q||p)$ is always non-negative and zero if and only if ${q=p}$, we reach the minimum exactly for ${q=p}$.
Obviously, optimizing over all possible probability distributions ${q\in \mathbb{M}}$ is generally intractable. 
Based on some general assumption, free energy methods simplify the problem in~\eqref{eq:prob_inference} to the minimization of a variational free energy term. We refer the reader to \citep{yedidia_constructing_2005} for a detailed elaboration on this topic.

The Bethe approximation restricts the distribution $q(\mathcal{X})$ to be a product of univariate distributions $b_n(x_n)$ and joint distributions $b_{n,m}(x_n,x_m)$ between pairs ${(n,m) \in \mathcal{E}}$:
\begin{equation*}
    q_\text{Bethe}\left( \mathcal{X} \right) := \prod_{n=1}^N b_n(x_n) \prod_{\mathclap{(n,m)\in \mathcal{E}}} b_{n,m}(x_n,x_m). \label{eq:bethe_approx}
\end{equation*}
This simplification leads to the \emph{Bethe free energy}
\begin{align*}
    F_\text{Bethe} = &\sum_{\mathclap{(n,m)\in \mathcal{E}}} \quad \sum_{{x_n,x_m}} b_{n,m}(x_n,x_m) \log \mleft( \frac{b_{n,m}(x_n,x_m)}{\phi_{n,m}(x_n,x_m)}\mright) \\
    &-\sum_{n=1}^N (|\mathcal{X}_n| - 1) \sum_{x_n} b_n(x_n) \log \mleft( \frac{b_n(x_n)}{\psi_n(x_n)} \mright),
\end{align*}
with ${\phi_{n,m}(x_n,x_m)} := {\psi_n(x_n) \psi_{n,m}(x_n,x_m) \psi_m(x_m)}$.
Moreover, the Bethe approximation relaxes the search space in~\eqref{eq:prob_inference} from the marginal polytope $\mathbb{M}$ to the \emph{local polytope} 
\begin{align*}
    \mathbb{L} = \Big \{ \hspace{-2pt} \{&b_n(x_n), b_{n,m}(x_n,x_m): \forall n \in [1;N], (n,m) \in \mathcal{E} \} \hspace{-2pt} :  \\
    & \sum_{x_n} b_{n,m}(x_n,x_m) = b_m(x_m), \, \sum_{x_n} b_n(x_n) = 1 \\
    & \sum_{x_m} b_{n,m}(x_n,x_m) = b_n(x_n)\Big \}.
\end{align*}
This means that the distributions ${b_n(x_n)}$ and ${b_{n,m}(x_n,x_m)}$ only need to locally fulfill consistency in a pairwise sense.
In summary, the Bethe approximation converts~\eqref{eq:prob_inference} into the optimization problem
\begin{equation}
    q_\text{Bethe} = \argmin2_{{\left\{b_n, b_{n,m} \right\} \in \mathbb{L}}} \, F_\text{Bethe} \mleft( \left\{b_n, b_{n,m} \right\} \mright). \label{eq:Bethe_minimization}
\end{equation}

\citet{yedidia_generalized_2000} showed that the fixed points of \ac{BP} applied to a factor graph correspond to the stationary points of the respective Bethe free energy. Seen in this light, \ac{BP} is a suboptimal algorithm to minimize~$F_\text{Bethe}$. The approximative nature in this sense is twofold:
First, there may exist multiple fixed points of the \ac{SPA} for the same factor graph, i.e., the solution of the (converged) \ac{BP} might correspond to an extremum of~$F_\text{Bethe}$ other than the global minimum in~$\mathbb{L}$~~\citep{knoll_fixed_2018}.
Second, the beliefs only fulfill the pairwise consistency constraints at the fixed points of \ac{BP}. This means that the solution only lies within the local polytope $\mathbb{L}$ after \ac{BP} has converged. However, \ac{BP} does not necessarily converge and failure of convergence is a major error mode~\citep{yuille_cccp_2002}. 

Various methods to directly solve \eqref{eq:Bethe_minimization} or variants thereof were proposed (see~\citep{yedidia_constructing_2005} and references therein).
\citet{yuille_cccp_2002} proposed to decompose the Bethe free energy into concave and convex parts which enables the application of a \ac{CCCP}. That algorithm consists of a double loop where the outer loop iteratively minimizes $F_\text{Bethe}$ and the inner loop ensures that the pairwise consistency constraints are fulfilled. Due to the \ac{CCCP}, the algorithm provably converges to an extremum of the Bethe free energy. 

\subsection{Examples}\label{sec:examples}
For the remainder of this section, we introduce two important classes of factor graphs which are the basis for the numerical experiments in Sec.~\ref{sec:experiments}. 
\paragraph{Example~1 - Ising Graphs}\label{example:Ising}
We consider factor graphs with ${N=M^2}$ variable nodes, arranged in a square 2D lattice, in which pairs of adjacent variable nodes $(x_n,x_m)$ are symmetrically coupled by the weights~$J_{n,m}$ via factor nodes~${\psi_{n,m}(x_n,x_m)} = {\exp \mleft( J_{n,m} x_n x_m \mright)}$.
Additionally, each variable node $x_n$ has local evidence in the form of a degree-1 factor node ${\psi_n(x_n) = \exp \left( \theta_n x_n \right)}$.
The Ising model originates from statistical physics where the binary variables ${x_n \in \{+1,-1\}}$ represent the orientation of elementary magnets in a lattice~\citep{peierls_isings_1936}. Each magnet is exposed to a local field~$\theta_n$ and is influenced by its neighbors via an assigned pairwise coupling~$J_{n,m}$.
Besides its fundamental significance in statistical physics, the Ising model is a universal mathematical model and finds applications in many other scientific domains such as image processing~\citep{besag_statistical_1986} and modeling of social networks~\citep{banerjee_model_2008,wainwright_graphical_2008}.

Following~\citep{yedidia_constructing_2005, mooij_sufficient_2007, knoll_fixed_2018}, we study the fully connected $2 \times 2$ Ising model, i.e., ${M=2}$ and ${N=4}$, where every pair of variable nodes is connected.
A factor graph representation of this model is given in Fig.~\ref{fig:Ising2x2}. 
With more cycles than variable nodes and a girth of~$3$, this graph can be parametrized to a highly frustrated system and is thus able to highlight the weaknesses of the \ac{SPA}~\citep{yedidia_constructing_2005}. In particular, we consider the Ising spin glass, where the parameters $\theta_n$ and $J_{n,m}$ are \ac{iid} random variables, sampled from a uniform distribution $\mathcal{U}[-S,+S]$ with ${S\in\mathbb{R}^+}$.
We are interested in the computation of the marginal functions
\begin{equation}
    f(x_n) = \sum_{\sim \{x_n\}} f(x_1,x_2,x_3,x_4), \quad n=1,2,3,4, \label{eq:Ising_APP}
\end{equation}
which correspond to marginal probability distributions ${p(x_n)=f(x_n)}$ if the Ising graph represents a probabilistic model.
While the direct computation of \eqref{eq:Ising_APP} is still feasible for our example with ${N=4}$, the number of summations grows exponentially with $N$, which calls for alternative methods with lower complexity. Applying the \ac{SPA} on the factor graph in Fig.~\ref{fig:Ising2x2} yields the single beliefs $b_n(x_n)$ as an approximation of $f(x_n)$ with a complexity that only grows quadratically with $N$.

\paragraph{Example 2 - Symbol Detection}\label{example:symbol_detection}
We study the problem of symbol detection in a digital communication system~\citep{proakis_digital_2007}. A transmitter sends a sequence of $N$ independent and uniformly distributed symbols ${c_n \in \{+1,-1\}}$ over a linear channel with memory, impaired by \ac{AWGN}. The receiver observes the sequence
\begin{equation}
    \vek{y} = 
    \underbrace{
    \begin{pmatrix} 
    h_0   &     &  &  \\
    \vdots& h_0  & \scalebox{1.3}{$\vek{0}$} &  \\
    h_L   &\vdots& \ddots & \\
         & h_L  &        & h_0\\
     &   \scalebox{1.3}{$\vek{0}$}  & \ddots  & \vdots \\
         &      &  & h_L\\
    \end{pmatrix}
    }_{=: \vek{H}}
    \underbrace{
    \begin{pmatrix}
        c_1 \\ c_2 \\ \vdots \\ \\ c_N
    \end{pmatrix}
    }_{=: \vek{c}}
    + 
    \underbrace{
    \begin{pmatrix}
        w_1 \\ w_2 \\ \vdots \\ \\ w_{N+L}
    \end{pmatrix}
    }_{=: \vek{w}}
    ,
\end{equation}
where ${\vek{h} \in \mathbb{R}^{L+1}}$ describes the impulse response of the channel of length ${L+1}$ and ${w_k \sim \mathcal{CN}(0,\sigma^2)}$ are independent noise samples from a complex circular Gaussian distribution.
Applying Bayes' theorem, the posterior distribution $p(\vek{c}|\vek{y})$
can be expressed in terms of the likelihood:
\begin{equation*}
    p(\vek{c}|\vek{y}) = \frac{1}{Z} p(\vek{y}|\vek{c}) = \frac{1}{Z} \exp \mleft ( -\frac{\left( \vek{y-Hc} \right) ^2}{\sigma^2} \mright ).
\end{equation*}
In the context of symbol detection, we want to infer the transmit symbols $c_n$ based on the channel observation $\vek{y}$, i.e., we are interested in the marginal distributions $p(c_n|\vek{y})$. 
Based on an observation model by \citet{ungerboeck_adaptive_1974}
\begin{equation*}
     p(\vek{y}|\vek{c}) \propto \exp\mleft( {2{\text{Re}}\left\{ \bm{c}^{\textrm H} \bm{H}^{\textrm H} \bm{y} \right\} -\bm{c}^{\textrm H}\bm{H}^{\textrm H}\bm{H}\bm{c}\over \sigma^2}  \mright),
\end{equation*}
we can factorize the likelihood
\begin{equation}
    p(\vek{y}|\vek{c}) = \frac1Z \prod\limits_{n=1}^{N} \left [ {F_n(c_n) }
                \prod\limits_{\substack{m=1 \\ m < n}}^N  {I_{n,m}(c_n,c_m) } \right ]
\label{eq:Ungerboeck_factorization}
\end{equation}
into the factors
\begin{align*}
    F_n(c_n) &:=  \exp \mleft( \frac{1}{\sigma^2} {\text{Re}} \mleft\{ 2 x_n c_n^\star - G_{n,n}|c_n|^2 \mright\} \mright) \label{eq:f_fn} \\
    I_{n,m}(c_n,c_m) &:= \exp \mleft(  -\frac{2}{\sigma^2} {\text{Re}} \mleft\{  G_{n,m} c_m c_n^\star  \mright\}  \mright),
\end{align*}
where ${\vek{x}:=\vek{H}^\text{H} \vek{y}}$ and ${\vek{G}:=\vek{H}^\text{H}\vek{H}}$ are the matched filtered versions of the observation and the channel matrix, respectively.
Modeling a factor graph based on~\eqref{eq:Ungerboeck_factorization} and applying the \ac{SPA} yields a low-complexity symbol detection algorithm, originally proposed by~\citet{colavolpe_siso_2011}.

\section{Message Passing for Cyclic Graphs}\label{sec:nmp}
Despite its drawbacks on cyclic graphs, the amazing success of the \ac{SPA} lies in its simplicity and generality: it is only defined by a \emph{local} message update rule which can be applied to any generic factor graph based on a suitable message update schedule.
Driven by this elegant concept, we are interested in finding message passing algorithms that perform well on graphs with many cycles where the \ac{SPA} fails. 
More specifically, we ask the following questions: 
\begin{itemize}
    \item If the \ac{SPA} fails to converge, does an alternative \emph{local} message update rule exist that converges (possibly to an extremum of the Bethe free energy) and which provides better results than the \ac{SPA}?
    \item If the \ac{SPA} converges to an extremum of the Bethe free energy, is there a local message update rule which yields superior performance, either because the \ac{SPA} converges to a fixed point which only corresponds to a local instead of global minimum of the Bethe free energy, or because the Bethe approximation itself is a bad approximation in this case?
\end{itemize}

\subsection{On Message Update Rules}
A message update rule defines a mapping from one or multiple incident messages to one outgoing message, which is applied \emph{locally} at the variable or factor nodes of a factor graph. Besides the initialization of the messages and their update schedule, these mappings fully define a graph-based inference algorithm.
The \ac{SPA} update rule at the variable nodes~\eqref{eq:VN_update_generic} is simply the product of all extrinsic messages. We adopt this quite intuitive aggregation principle and focus on finding a message update rule for the factor nodes, i.e., an alternative to~\eqref{eq:FN_update_generic}.
For factor nodes of degree~2, such as in~\eqref{eq:mrf}, the update rule simplifies to a mapping from one single incident message to one outgoing message:
\begin{equation}
    \text{FN}_\text{e}(\psi_{n,m}): m_{x_n \rightarrow \psi_{n,m}}(x_n) \mapsto m_{\psi_{n,m} \rightarrow x_m}(x_m). \label{eq:extrinsic_update}
\end{equation}
If the pairwise factors $\psi_{n,m}(x_n,x_m)$ are symmetric with regard to $x_n$ and $x_m$, and follow the exponential form 
\begin{equation*}
    \psi_{n,m}(x_n,x_m) = \exp \mleft( E_{n,m} x_n x_m \mright), \; x_n,x_m \in \{+1,-1\},
\end{equation*}
we can distill the dependency from the function $\psi_{n,m}$ to the scalar parameter ${E_{n,m}\in\mathbb{R}}$, which quantifies the repulsive (${E_{n,m}<0}$) or attractive (${E_{n,m}>0}$) coupling between the nodes $x_n$ and $x_m$. This directly coincides with the pairwise coupling weights ${J_{n,m} = E_{n,m}}$ of the Ising model in Example~1. The factor nodes $I_{n,m}$ of Example~2 can be reduced to the coupling parameters ${E_{n,m} = -2G_{n,m} / \sigma^2}$.

\paragraph{Challenging the Extrinsic Principle}
Most of the existing message passing algorithms follow the extrinsic information principle. For instance in turbo decoding, it is known to be an important property of good message passing decoders~\citep{richardson_capacity_2001}. 
Ensuring that only extrinsic messages are received, it prevents backcoupling of intrinsic information in tree-structured graphs, which would otherwise lead to a self-enhancement of the messages, also known as ``double counting''. Thereby, it guarantees that the \ac{SPA} is exact on trees \citep{kuck_belief_2020}.
We argue that this is in general not valid for cyclic graphs where backcoupling of messages is inevitable due to the very nature of the cycles.
Therefore, we propose a second message update rule which operates contradictory to the extrinsic principle: instead of ignoring the intrinsic message, the message update should rather actively leverage this additional information, e.g., to ensure that local consistency between neighboring nodes is fulfilled.

Without the extrinsic principle, we need to reconsider the messages from degree-1 factor nodes which are then also subject to iterative updates. To avoid an increase in complexity due to additional message updates at the degree-1 factor nodes, we apply a clustering approach similar to~\citep{rapp_structural_2022}. We split up the single factors $\psi_n(x_n)$ into $|\mathcal{X}_n|$ parts ${\Psi_n(x_n) := \left(\psi_n(x_n)\right)^{\frac{1}{|\mathcal{X}_n|}}}$ and merge them into the adjacent pairwise factors $\psi_{n,m}(x_n,x_m)$, such that the new clustered factors are
\begin{equation*}
    \Psi_{n,m}(x_n,x_m) := \Psi_n\mleft(x_n\mright) \psi_{n,m}\mleft(x_n,x_m\mright) \Psi_m\mleft(x_m\mright).
\end{equation*}
The overall factorization~\eqref{eq:mrf} simplifies to
\begin{equation*}
    f(x_1,\ldots,x_N) = \frac1Z \prod_{{(n,m)\in \mathcal{E}}} \Psi_{n,m}(x_n,x_m),
\end{equation*}
which leads to the non-extrinsic mapping
\begin{equation}
    \text{FN}(\Psi_{n,m}): 
    \begin{pmatrix}
        m_{x_n \rightarrow \Psi_{n,m}}(x_n) \\ m_{x_m \rightarrow \Psi_{n,m}}(x_m)
    \end{pmatrix}
     \mapsto m_{\Psi_{n,m} \rightarrow x_m}(x_m). \label{eq:nonextrinsic_update}
\end{equation}
If the single factors are in exponential form 
\begin{equation*}
    \Psi_n(x_n) = \exp \left( E_n x_n \right), \quad x_n \in \{ +1,-1\},
\end{equation*}
the clustered factors $\Psi_{n,m}(x_n,x_m)$ are fully characterized by the three scalars $E_n, E_{n,m}$ and $E_m$.

\subsubsection{Neural Networks as Function Approximators}
Finding suitable mappings~\eqref{eq:extrinsic_update} or~\eqref{eq:nonextrinsic_update} such that the overall message passing algorithm performs well is generally non-trivial.
We employ feed-forward \acp{NN}, known to be efficient universal function approximators~\citep{hornik_multilayer_1989}, to reduce the search space of all possible mappings to a set of weights and biases~$\mathcal{P}$, which fully parametrize the \ac{NN}. At a factor node $f_j$, the network accepts $N_\text{in}$ inputs and produces the updated outgoing message $m_{f_j \rightarrow x_n}$.
For factor graphs with binary variables~$x_n$, the messages $m_{f_j \rightarrow x_n}(x_n)$ can be expressed in scalar \acp{LLR}
\begin{equation*}
    L_{f_j \rightarrow x_n} := \log \mleft( \frac{m_{f_j \rightarrow x_n}(x_n=+1)}{m_{f_j \rightarrow x_n}(x_n=-1)} \mright).
\end{equation*}
A similar definition holds for the \acp{LLR} $L_{x_n \rightarrow f_j}$ based on the messages $m_{x_n \rightarrow f_j}(x_n)$.
For the extrinsic update~\eqref{eq:extrinsic_update}, there are ${N_\text{in}=2}$ inputs: the \ac{LLR} of the incoming extrinsic message and the coupling parameter $E_{n,m}$ of the local factor node. Without the extrinsic principle, the \ac{NN} furthermore accepts the \ac{LLR} of the intrinsic message as well as $E_n$ and $E_m$, i.e., in total $N_\text{in}=5$ inputs.
Since we only approximate a local mapping from a few scalar inputs to a single output, we can choose a very compact \ac{NN} structure with a single hidden layer and $7$ neurons, as summarized in Table~\ref{tab:NN_architecture}. 
\begin{table}[tb]
    \centering
\caption{\ac{NN} Architecture}
\begin{tabular}{ l l r }
  \toprule
    Layer (linear) & Activation & Dimension \\
  \midrule			
    Input   & ReLU   & $(N_\text{in},7)$ \\
    Hidden  & Tanh   & $(7,7)$ \\
    Output  & Linear & $(7,1)$ \\
  \bottomrule  
\end{tabular}
\label{tab:NN_architecture}
\end{table}

Having set up the \ac{NN} structure, we are able to define a convenient message update rule by appropriately tuning the parameterization $\mathcal{P}$ of the \ac{NN}.
We are interested in a local update rule such that the overall message passing performs well. To this end, we optimize $\mathcal{P}$ with respect to an objective function that evaluates the end-to-end performance of the inference task.
Therefore we apply a fixed number of message passing iterations and back-propagate the gradient of the objective function in order to iteratively optimize $\mathcal{P}$ using gradient descent based on a representative set of examples. 
Note that this data-driven approach inevitably leads to a specialization of the learned message update to the data. However, we expect the result to be fairly generic and to have good generalization capabilities since we only optimize very few parameters in an otherwise model-aware system.
Moreover, despite the end-to-end optimization, we only use a single message update rule for the entire factor graph, i.e., we employ the same instance of the \ac{NN} for the message updates at all factor nodes and in each iteration\footnote{As a consequence, the training procedure of the \ac{NN} is not entirely local because the local copies of the \ac{NN} at each factor node must be globally synchronized during optimization. However, the local nature of the message updates is still retained.}.

We note that our approach can be interpreted as a special instance of a \ac{GNN} as, e.g., described by \citet{yoon_inference_2019}.
In comparison, our model passes scalar messages instead of high-dimensional vectors and does not use any hidden states or embeddings at the variable nodes. For this reason, we do not require a second \ac{NN} with a gated recurrent unit, as used in \citep{yoon_inference_2019} to update the hidden states based on the aggregated messages. Furthermore, we do not require a third \ac{NN} which implements a trainable readout function to interpret the final node embeddings. 

\subsection{End-to-end Objective Functions}\label{sec:loss}
In the generic context of marginal inference, we hope to find a good approximation of the true marginals. A convenient objective function is the \ac{KL} divergence which measures a type of statistical distance between the beliefs $b_n(x_n)$ and the exact marginal distributions ${p(x_n) = \sum_{\sim\{x_n\}} p(x_1,\ldots,x_N)}$:
\begin{equation}
    \mathcal{L}_\text{KL} := D_\text{KL}\left(b_n(x_n) \middle\| p(x_n) \right). \label{eq:KL_to_APP}
\end{equation}
For large graphs, the computation of $p(x_n)$ might be infeasible, and $\mathcal{L}_\text{KL}$ becomes impractical. Therefore, we propose alternative loss functions in what follows.

The training of a symbol detector as in Example~2 is a typical supervised learning scenario where the labels are given by the transmitted symbols $c_n$. An appropriate performance measure for symbol detection is the \ac{BMI}
which is an achievable information rate\footnote{In our case, where the symbols $c_n$ follow a Rademacher distribution, the \ac{BMI} is equivalent to the mutual information.} for our scenario~\citep{Fabregas_foundations_2008}. 
By a sample mean estimate over $D$ labeled examples ${(\vek{c},\vek{y})}$ from the data batch $\mathcal{D}$, the \ac{BMI} can be approximated by
\begin{equation*}\label{bmi_est}
    \text{BMI} \approx 1 -
    \frac{1}{D N} 
    \sum\limits_{n=1}^{N}
    \sum\limits_{(\vek{c}, \vek{y})\in \mathcal{D}}
    \log_2 \mleft (  \text{e}^{-c_n L_{n}(\vek{y})} + 1  \mright ),
\end{equation*}
where ${L_n(\vek{y})}$ denotes the \ac{LLR} from the belief ${b_n(c_n)}$~\citep{alvarado_achievable_2018}.

Other applications such as the Ising model in Example~1 relate to the class of unsupervised problems if the true marginals are not accessible.
For such scenarios, we consider a novel and application-agnostic objective function in the following. Inspired by the Bethe approximation, which is known to yield excellent results for many applications, even in cases where the \ac{SPA} performs poorly~\citep{yuille_cccp_2002}, we propose a regularized minimization of the Bethe free energy:
\begin{equation}
    \mathcal{L}_\text{Bethe} := F_\text{Bethe} + \alpha \mathcal{L}_{\mathbb{L}}, \quad \alpha \in \mathbb{R}^+.
    \label{eq:constrainedBethe}
\end{equation}
To ensure local consistency, we introduce the \emph{Bethe consistency distance}
\begin{align*}
    \mathcal{L}_{\mathbb{L}} := \; &D_\text{KL}\left( \sum_{x_m} b_{n,m}(x_n,x_m) \middle\| b_n(x_n) \right) \nonumber \\ 
    &+ D_\text{KL}\left( \sum_{x_n} b_{n,m}(x_n,x_m) \middle\| b_m(x_m) \right)
\end{align*}
as a type of distance measure between the solution of the approximative inference ${\{ b_n,b_{n,m} \}}$ and the local polytope~$\mathbb{L}$. 
The weight $\alpha$ in~\eqref{eq:constrainedBethe}  is a hyperparameter that controls how strictly the local consistency is enforced. With this penalty term~$\mathcal{L}_\mathbb{L}$, we hope to suppress oscillations in the message passing, as they occur in the \ac{SPA} for graphs with strong coupling.

\section{Experiments}\label{sec:experiments}
We consider the examples of Sec.~\ref{sec:examples} for numerical evaluation. To enable a deeper analysis, we fix the number of variable nodes to ${N=4}$ such that the computation of the true marginals is feasible. Despite this rather small extent, these models lead to factor graphs with a high density of short cycles and are thus expressive examples to highlight the weaknesses of the \ac{SPA}. 
Furthermore, we fix the global settings of the message passing to standard choices: all \ac{LLR} messages are initialized with zero and we perform 10 iterations of a parallel schedule, i.e., each iteration comprises the parallel update of all messages at the factor nodes followed by message updates at all variable nodes.

A common technique to improve the performance of the \ac{SPA} on graphs with cycles is the use of ``momentum'', i.e., replacing a message $L^{(t)}$ of the \ac{SPA} in iteration $t$ with the weighted average $(1-\mu) L^{(t)} + \mu L^{(t-1)}$ \citep{murphy_loopy_1999}. By choosing ${0< \mu < 1}$, the idea is to improve the convergence behavior of the message passing scheme compared to the original \ac{SPA} (${\mu=0}$) while retaining the same fixed points. As in~\citep{murphy_loopy_1999}, we set ${\mu=0.1}$ and use this variant of the \ac{SPA} as an additional baseline in the following experiments, where we refer to it as \ac{SPA}$_\mu$.

Besides the \ac{SPA}, we similarly apply message passing based on the newly proposed update rule~\eqref{eq:nonextrinsic_update}. We call the resulting inference algorithm cycBP (\ac{BP} for cyclic graphs). If we use the extrinsic update rule~\eqref{eq:extrinsic_update}, we denote the algorithm with $\text{cycBP}_\text{e}$. We also consider the CCCP for the Bethe free energy as defined in~\citep{yuille_cccp_2002}, since it gives interesting insights into the quality of the Bethe approximation. For the double loop, we apply $25$~outer iterations, each comprising $25$~inner iterations. 

\paragraph{Ising model}\label{sec:experiments_Ising}
We study the $2 \times 2$ fully connected spin glass model of Example~1 for ${S=2}$, i.e., all parameters $\theta_n$ and $J_{n,m}$ are independently sampled from a uniform distribution $\mathcal{U}[-2,+2]$. Table~\ref{tab:results_spinGlass} evaluates the behavior of all discussed inference schemes, averaged over $10^5$ different graphs. $\hat{\sigma}_{\mathcal{L}_\text{KL}}$ denotes the empirical standard deviation of ${\mathcal{L}}_\text{KL}$ of the individual graphs from the empirical mean.
\begin{table}[tb]
    \centering
\caption{Behavior of the Novel Message Passing Algorithm cycBP for the $2 \times 2$ Spin Glass, Averaged over $10^5$ Graphs}
\begin{tabular}{ l l c c c l}
  \toprule
    Algo. & Loss & ${\mathcal{L}}_\text{KL}$ & $\hat{\sigma}_{\mathcal{L}_\text{KL}}$ & $F_\text{Bethe}$ & $\mathcal{L}_\mathbb{L}$ \\
  \midrule		
    \ac{SPA} & \phantom{0.}- & $0.087$ & $0.265$ & $-7.50$ & $0.30$ \\ 
    \ac{SPA}$_{\mu}$ & \phantom{0.}- & $0.035$ & $0.113$ & $-7.49$ & $0.12$ \\
    CCCP & \phantom{0.}- & $0.044$ & $0.094$ & $-7.24$ & $2 \cdot 10^{-6}$ \\
    $\text{cycBP}_\text{e}$ & $\mathcal{L}_\text{KL}$ & $0.040$ & $0.068$ & $-7.37$ & $0.17$ \\
    cycBP & $\mathcal{L}_\text{KL}$ & $0.014$ & $0.023$ & $-7.37$ & $0.48$ \\
    $\text{cycBP}_\text{e}$ & $\mathcal{L}_\text{Bethe}$ & $0.030$ & $0.054$ & $-7.38$ & $0.11$ \\
    cycBP & $\mathcal{L}_\text{Bethe}$ & $0.027$ & $0.057$ & $-7.47$ & $0.027$ \\
  \bottomrule  
\end{tabular}
\label{tab:results_spinGlass}
\end{table}
We can observe that the \ac{SPA} does not leverage the full potential of the Bethe approximation, since the average loss ${\mathcal{L}_\text{KL}=0.087}$ of the \ac{SPA} is twice as large compared to ${\mathcal{L}_\text{KL}=0.044}$ for the CCCP. Although the \ac{SPA} reaches on average a smaller $F_\text{Bethe}$ than the CCCP, the beliefs of the \ac{SPA} show local inconsistencies with~${\mathcal{L}_\mathbb{L}=0.3}$ due to non-convergent behavior. Using ``momentum'' in the \ac{SPA} message updates can help to mitigate this behavior: the \ac{SPA}$_\mu$ shows improved pairwise consistency~${\mathcal{L}_\mathbb{L}=0.12}$ and also yields in average a better approximation of the true marginals~(${\mathcal{L}_\text{KL}=0.035}$).
The CCCP has a vanishing Bethe consistency distance $\mathcal{L}_\mathbb{L}$, i.e., the results of the \ac{CCCP} lie within the local polytope~$\mathbb{L}$.
We search for alternative message update rules, by optimizing $\mathcal{P}$ of the \ac{NN}-based mappings towards minimal~$\mathcal{L}_\text{KL}$. The training batches are sampled from a spin glass model with ${S=3}$ to put more emphasis on graphs with strong coupling, where the \ac{SPA} is known to be susceptible to convergence errors. The results in Tab.~\ref{tab:results_spinGlass} show that there indeed exist superior message update rules to the \ac{SPA} for this class of cyclic graphs. Using the extrinsic update rule~\eqref{eq:extrinsic_update}, the $\text{cycBP}_\text{e}$ algorithm reaches ${\mathcal{L}_\text{KL}=0.04}$ and thereby outperforms the original \ac{SPA} as well as the \ac{CCCP}.

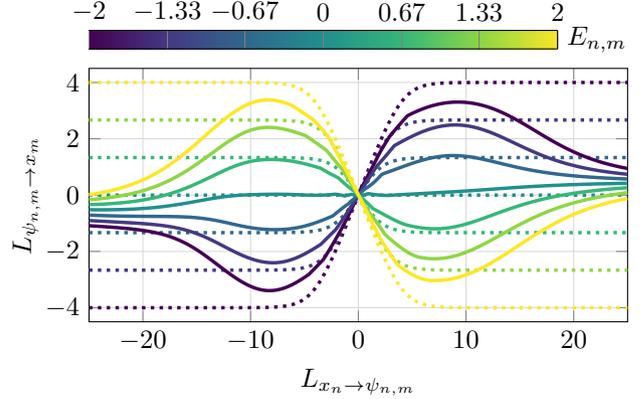
\begin{figure}[t]
\centering
\pgfplotsset{compat=newest}
    \begin{tikzpicture}[baseline, trim axis right]
    \begin{axis}[
          width=1.05\linewidth, %
          height=0.6\linewidth,
          grid=major, %
          grid style={gray!30}, %
          xlabel= $L_{x_n \rightarrow \psi_{n,m}}$,
          ylabel= $L_{\psi_{n,m} \rightarrow x_m}$,
          y label style={yshift=-5pt},
          enlarge x limits=false,
          enlarge y limits=false,
          xmin = -25,
          xmax = +25,
          ymin = -4.5,
          ymax = +4.5,
          point meta min=-2,
          point meta max=2,
          colormap={c_color}{samples of colormap={7 of viridis}},
          colorbar,
          colorbar horizontal,
          colorbar style={height=7pt, at={(0,1.15)},anchor=north west, xtick={-2,-1.33,-0.67,0.0,0.67,1.33,2},xticklabel pos=upper,ylabel=$E_{n,m}$,
          axis y line*=right,
          ylabel style={rotate=-90},
          width=.87 *\pgfkeysvalueof{/pgfplots/parent axis width},
          },
        ]   
        
        \addplot[index of colormap=0, line width=1.3pt, dotted] table[x=LLR_in, y=LLR_out_BP J0 ,col sep=comma] {num_results/Ising/table_2x2Ising_LLR_in_out.csv}; 
        \addplot[index of colormap=1, line width=1.3pt, dotted] table[x=LLR_in, y=LLR_out_BP J1 ,col sep=comma] {num_results/Ising/table_2x2Ising_LLR_in_out.csv}; 
        \addplot[index of colormap=2, line width=1.3pt, dotted] table[x=LLR_in, y=LLR_out_BP J2 ,col sep=comma] {num_results/Ising/table_2x2Ising_LLR_in_out.csv}; 
        \addplot[index of colormap=3, line width=1.3pt, dotted] table[x=LLR_in, y=LLR_out_BP J3 ,col sep=comma] {num_results/Ising/table_2x2Ising_LLR_in_out.csv}; 
        \addplot[index of colormap=4, line width=1.3pt, dotted] table[x=LLR_in, y=LLR_out_BP J4 ,col sep=comma] {num_results/Ising/table_2x2Ising_LLR_in_out.csv};
        \addplot[index of colormap=5, line width=1.3pt, dotted] table[x=LLR_in, y=LLR_out_BP J5 ,col sep=comma] {num_results/Ising/table_2x2Ising_LLR_in_out.csv};
        \addplot[index of colormap=6, line width=1.3pt, dotted] table[x=LLR_in, y=LLR_out_BP J6 ,col sep=comma] {num_results/Ising/table_2x2Ising_LLR_in_out.csv};
        \addplot[index of colormap=0, line width=1.3pt] table[x=LLR_in, y=LLR_out_NMP J0 ,col sep=comma] {num_results/Ising/table_2x2Ising_LLR_in_out.csv}; 
        \addplot[index of colormap=1, line width=1.3pt] table[x=LLR_in, y=LLR_out_NMP J1 ,col sep=comma] {num_results/Ising/table_2x2Ising_LLR_in_out.csv}; 
        \addplot[index of colormap=2, line width=1.3pt] table[x=LLR_in, y=LLR_out_NMP J2 ,col sep=comma] {num_results/Ising/table_2x2Ising_LLR_in_out.csv}; 
        \addplot[index of colormap=3, line width=1.3pt] table[x=LLR_in, y=LLR_out_NMP J3 ,col sep=comma] {num_results/Ising/table_2x2Ising_LLR_in_out.csv}; 
        \addplot[index of colormap=4, line width=1.3pt] table[x=LLR_in, y=LLR_out_NMP J4 ,col sep=comma] {num_results/Ising/table_2x2Ising_LLR_in_out.csv};
        \addplot[index of colormap=5, line width=1.3pt] table[x=LLR_in, y=LLR_out_NMP J5 ,col sep=comma] {num_results/Ising/table_2x2Ising_LLR_in_out.csv};
        \addplot[index of colormap=6, line width=1.3pt] table[x=LLR_in, y=LLR_out_NMP J6 ,col sep=comma] {num_results/Ising/table_2x2Ising_LLR_in_out.csv};

    \end{axis}
\end{tikzpicture}
    \caption{The mapping \eqref{eq:extrinsic_update} of the $\text{cycBP}_\text{e}$ algorithm, optimized for the $2 \times 2$ spin glass with ${S=3}$ (solid), in comparison with the mapping of the \ac{SPA} (dotted), plotted for different coupling $E_{n,m}$.}
    \label{fig:LLR_in_out}
\end{figure}

We visualize the message update rule of the $\text{cycBP}_\text{e}$ algorithm in Fig.~\ref{fig:LLR_in_out} by plotting the optimized mapping~\eqref{eq:extrinsic_update} from the incoming \ac{LLR} message $L_{x_n \rightarrow \psi_{n,m}}$ to the outgoing \ac{LLR} message $L_{\psi_{n,m} \rightarrow x_m}$. Similar to the \ac{SPA}, the mapping is point-symmetric to the origin. The major difference is the behavior for incident \ac{LLR} messages with high magnitudes ${|L_{x_n \rightarrow \psi_{n,m}}|>8}$, where the outgoing messages are heavily attenuated. Intuitively, this behavior reduces the potential of oscillation in graphs with strong coupling~$E_{n,m}$.
We can further improve the inference performance by disabling the extrinsic principle in the message passing procedure. The resulting algorithm cycBP can be interpreted as a generalization of $\text{cycBP}_\text{e}$ and outperforms the latter with ${\mathcal{L}_\text{KL}=0.014}$, as reported in Tab.~\ref{tab:results_spinGlass}. It also yields a superior approximation of the true marginals compared to the momentum-based \ac{SPA}$_\mu$, although the Bethe consistency distance~${\mathcal{L}_\mathbb{L}=0.48}$ is relatively high in this case.

Moreover, we consider unsupervised training towards the proposed loss function $\mathcal{L}_\text{Bethe}$. For the $\text{cycBP}_\text{e}$ algorithm, the unsupervised training leads to a smaller loss ${\mathcal{L}_\text{KL}=0.03}$ compared to the supervised training, i.e., the loss function $\mathcal{L}_\text{Bethe}$ is better suited for the optimization via stochastic gradient descent than the loss function $\mathcal{L}_\text{KL}$ in this case.
In the unsupervised training towards $\mathcal{L}_\text{Bethe}$, we observe substantial differences between the two variants $\text{cycBP}_\text{e}$ and cycBP: while the optimization of $\text{cycBP}_\text{e}$ converges reliably, the training of cycBP is unstable and the optimization needs to run multiple times with different initializations for $\mathcal{P}$ until a reasonable result is obtained. This behavior is also reflected in the results in Tab.~\ref{tab:results_spinGlass}, where the cycBP algorithm shows a degraded performance with ${\mathcal{L}_\text{KL}=0.027}$, compared to the supervised training (${\mathcal{L}_\text{KL}=0.014}$). We conjecture that this is accounted for by the local consistency constraint~$\mathcal{L}_\mathbb{L}$, which can be directly enforced at the message update at the factor nodes if the intrinsic message also takes part in the update. Optimization of the hyperparameter $\alpha$ did not lead to considerable changes in this behavior and we used ${\alpha=25}$ for all presented results.

\newcommand\markersize{0.95}
\newcommand\scatterborderx{0.06}
\newcommand\scatterbordery{0.07}
\newcommand\magIsingSize{0.61\linewidth}
\begin{figure}[t]
\pgfplotsset{colormap/jet}
\centering
\begin{tikzpicture}[trim axis right] %
    \begin{axis}
    [name=SPA,
    width=\magIsingSize,
    height=\magIsingSize,
    view={0}{90},
    xmin=-2-\scatterborderx,
    xmax=2+\scatterborderx,
    ymin=-2-\scatterbordery,
    ymax=2+\scatterbordery,
    ylabel=$J$,
    ylabel shift = -10 pt,
    ylabel style={rotate=-90},
    xticklabel style=white,
    xlabel shift = 10pt,
    colorbar,
    colorbar horizontal,
    colorbar style={
    width=1.93*\pgfkeysvalueof{/pgfplots/parent axis width},
    height = 7 pt,
    at={(0,1.15)},anchor=north west,
    ylabel=$\mathcal{L}_\text{KL}$,
    ylabel style={rotate=-90},
    xticklabel pos=upper,    
    axis y line*=right,
    },
    ]
        \addplot3
        [surf, 
        point meta min=0, point meta max=1.05, 
        scatter, 
        only marks, mark=square*, mark options={scale=\markersize,solid},
        ] 
        table [col sep=comma, x=theta, y=J, z=KL] {num_results/Ising/table_J_over_theta_BP_ref.csv};
    \end{axis}
    
    \begin{axis}
    [name=CCCP,
    at=(SPA.right of south east), anchor=left of south west,
    width=\magIsingSize,
    height=\magIsingSize,
    view={0}{90},
    xmin=-2-\scatterborderx,
    xmax=2+\scatterborderx,
    ymin=-2-\scatterbordery,
    ymax=2+\scatterbordery,
    ylabel style={rotate=-90},
    xticklabel style=white,
    xlabel shift = 10pt,
    yticklabels={,,},
    ]
        \addplot3
        [surf, 
        point meta min=0, point meta max=1.05, 
        scatter, 
        only marks, mark=square*, mark options={scale=\markersize,solid}] 
        table [col sep=comma, x=theta, y=J, z=KL] {num_results/Ising/table_J_over_theta_CCCP2_ref.csv};
    \end{axis}

    \begin{axis}
    [name=NMP_KL,
    at=(SPA.below south west), anchor=above north west,
    width=\magIsingSize,
    height=\magIsingSize,
    view={0}{90},
    xmin=-2-\scatterborderx,
    xmax=2+\scatterborderx,
    ymin=-2-\scatterbordery,
    ymax=2+\scatterbordery,
    ylabel style={rotate=-90},
    ylabel=$J$,
    ylabel shift = -10 pt,
    xlabel=$\theta$,
    ]
        \addplot3
        [surf, 
        point meta min=0, point meta max=1.05, 
        scatter, 
        only marks, mark=square*, mark options={scale=\markersize,solid}] 
        table [col sep=comma, x=theta, y=J, z=KL] {num_results/Ising/table_J_over_theta_trainedKL_NONextrinsic_feedDGR1msgs.csv};
    \end{axis}

    \begin{axis}
    [name=NMP_Bethe,
    at=(NMP_KL.right of south east), anchor=left of south west,
    width=\magIsingSize,
    height=\magIsingSize,
    view={0}{90},
    xmin=-2-\scatterborderx,
    xmax=2+\scatterborderx,
    ymin=-2-\scatterbordery,
    ymax=2+\scatterbordery,
    ylabel style={rotate=-90},
    xlabel=$\theta$,
    yticklabels={,,},
    ]
        \addplot3
        [surf, 
        point meta min=0, point meta max=1.05, 
        scatter, 
        only marks, mark=square*, mark options={scale=\markersize,solid}] 
        table [col sep=comma, x=theta, y=J, z=KL] {num_results/Ising/table_J_over_theta_trainedBethe_extrinsic.csv};
    \end{axis}

\end{tikzpicture}
    \caption{Approximation error $\mathcal{L}_\text{KL}$ on the $2 \times 2$ Ising model with constant parameters $\theta$ and $J$: \ac{SPA} (top left), CCCP (top right), cycBP optimized on spin glasses w.r.t. $\mathcal{L}_\text{KL}$ (bottom left), and $\text{cycBP}_\text{e}$ trained with $\mathcal{L}_\text{Bethe}$ (bottom right).}
    \label{fig:Ising_heatmaps}
\end{figure}
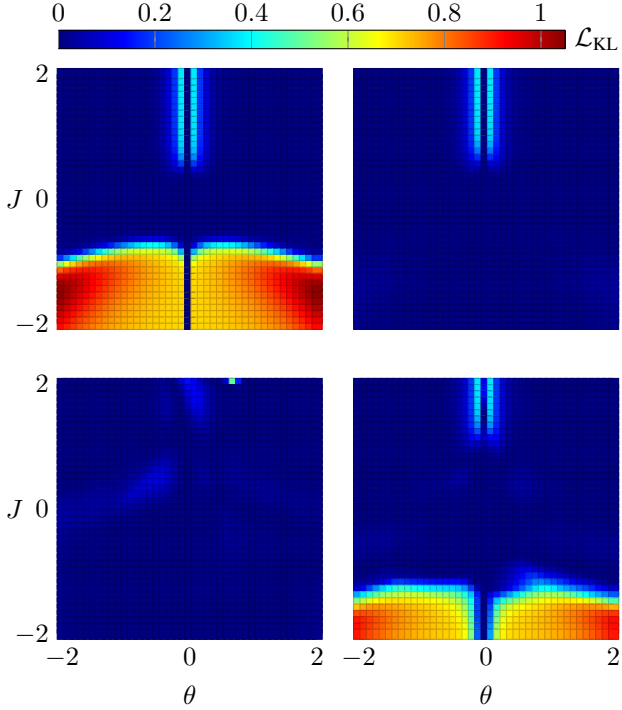
To analyze the convergence properties on highly frustrated systems, we consider the Ising model with constant parameters $\theta$ and $J$. Note that we do not specifically optimize the models for this scenario, but rather use the previous parametrization~$\mathcal{P}$ which is optimized for spin glasses with ${S=3}$. Figure~\ref{fig:Ising_heatmaps} plots $\mathcal{L}_\text{KL}$ over $\theta$ and $J$ for different inference algorithms. \citet{knoll_fixed_2018} showed that the Bethe free energy has a unique minimum in the complete antiferromagnetic domain (${J<0}$) and in large parts of the ferromagnetic case (${J>0}$), except for a region around ${\theta=0}$, where $F_\text{Bethe}$ has two  minima in $\mathbb{L}$. This coincides with our findings of the approximation error $\mathcal{L}_\text{KL}$ for the CCCP in Fig.~\ref{fig:Ising_heatmaps}. The \ac{SPA} shows failure to converge in large parts of the antiferromagnetic region with ${J<-1}$, where it does not converge to the unique fixed point and produces large approximation errors. The extrinsic message passing scheme $\text{cycBP}_\text{e}$, optimized towards $\mathcal{L}_\text{Bethe}$, shows an improved behavior. However, in the antiferromagnetic case with strong repellings ($J<-1.5$), there are still considerable approximation errors. The non-extrinsic message passing algorithm cycBP shows good inference capabilities over the complete considered region if it is optimized towards $\mathcal{L}_\text{KL}$. The unsupervised training with respect to $\mathcal{L}_\text{Bethe}$ leads to a similar performance as the CCCP, however, the training procedure is again relatively unstable in this case.

\paragraph{Symbol Detection}
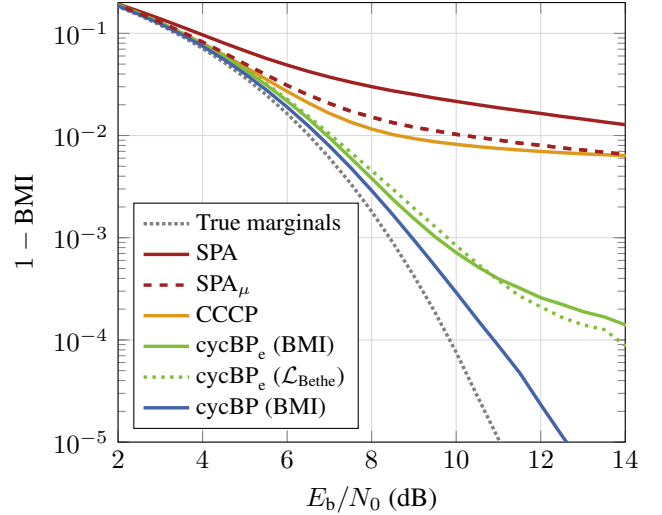
\begin{figure}[t]
\centering
    \begin{tikzpicture}[baseline, trim axis right]
    \begin{axis}[
          width=\linewidth, %
          height=0.9\linewidth,
          grid=major, %
          grid style={gray!30}, %
          xlabel= $E_\text{b}/N_0$ (dB),
          ylabel= ${1-\text{BMI}}$,
          ymode = log,
          enlarge x limits=false,
          enlarge y limits=false,
          xmin = 2,
          xmax = 14,
          ymin = 0.00001,
          ymax = 0.2,
          legend style={at={(.03,0.03)},anchor=south west, font=\small},
          legend cell align={left},
        ]
        \addplot[densely dotted, color=gray, line width=1.3pt] table[x=Eb/N0, y expr=1-\thisrowno{4},col sep=comma] {num_results/symbol_detection/BER+BMI_over_ebno_N4_BPSK_ref+nonextrinsic.csv}; 
        \addplot[color=KITred, line width=1.3pt] table[x=Eb/N0, y expr=1-\thisrowno{6},col sep=comma] {num_results/symbol_detection/BER+BMI_over_ebno_N4_BPSK_ref+nonextrinsic.csv}; 
        \addplot[dashed, color=KITred, line width=1.3pt] table[x=Eb/N0, y expr=1-\thisrowno{2},col sep=comma] {num_results/symbol_detection/BER+BMI_over_ebno_N4_BPSK_momentumBP.csv};
        \addplot[color=KITorange, line width=1.3pt] table[x=Eb/N0, y expr=1-\thisrowno{8} ,col sep=comma] {num_results/symbol_detection/BER+BMI_over_ebno_N4_BPSK_extrinsic.csv};
        \addplot[color=KITpalegreen, line width=1.3pt] table[x=Eb/N0, y expr=1-\thisrowno{2} ,col sep=comma] {num_results/symbol_detection/BER+BMI_over_ebno_N4_BPSK_extrinsic.csv}; 
        \addplot[dotted, color=KITpalegreen, line width=1.3pt] table[x=Eb/N0, y expr=1-\thisrowno{2} ,col sep=comma] {num_results/symbol_detection/BER+BMI_over_ebno_N4_BPSK_extrinsic_constrainedBethe.csv}; 
        \addplot[color=KITblue, line width=1.3pt] table[x=Eb/N0, y expr=1-\thisrowno{2} ,col sep=comma] {num_results/symbol_detection/BER+BMI_over_ebno_N4_BPSK_ref+nonextrinsic.csv}; 

        \addplot[dashed, color=KITred, line width=1.3pt] table[x=Eb/N0, y expr=1-\thisrowno{2},col sep=comma] {num_results/symbol_detection/BER+BMI_over_ebno_N4_BPSK_momentumBP.csv};
        
        \legend{True marginals \\ SPA \\ SPA$_\mu$ \\CCCP  \\  $\text{cycBP}_\text{e}$ ($\text{BMI}$)\\ $\text{cycBP}_\text{e}$ ($\mathcal{L}_\text{Bethe}$) \\ cycBP ($\text{BMI}$) \\}
    \end{axis}
\end{tikzpicture}
    \caption{Detection performance of the proposed cycBP algorithm, averaged over $10^7$ random channels.}
    \label{fig:BER_varChannel_N4}
\end{figure} 
We further consider the factor graphs of Example~2 for approximate symbol detection on linear channels with memory ${L=2}$.
To generate random channels, we independently sample each tap $h_\ell$ of the channel impulse response for every example from a Gaussian distribution with zero mean and unit variance and subsequently normalize each channel to unit energy ${\lVert \bm{h} \rVert_2 = 1}$.
Figure~\ref{fig:BER_varChannel_N4} evaluates the detection performance of the considered inference algorithms in terms of the \ac{BMI} over the signal-to-noise ratio ${E_\text{b}/N_0 = 1/\sigma^2}$. Both, the \ac{SPA} and the CCCP run into an error floor for high $E_\text{b}/N_0$, where the graphs tend to have strong coupling via the factor nodes $I_{n,m}(c_n,c_m)$. 
The momentum-based message updates of the \ac{SPA}$_\mu$ enhance the original \ac{SPA} in the entire ${E_\text{b}/N_0}$ range under consideration and close the performance gap to the \ac{CCCP}.
During the optimization of the new message update rules, the $E_\text{b}/N_0$ in dB was sampled from ${\mathcal{U}[0,16]}$ for each batch element independently.
To help the update rule adapt to different channel realizations, i.e., different $E_\text{b}/N_0$ and different channel taps $\bm{h}$, we feed $E_\text{b}/N_0$ and $\bm{h}$ as additional inputs to the \ac{NN}.
Figure~\ref{fig:BER_varChannel_N4} shows that the optimized algorithms $\text{cycBP}_\text{e}$ and cycBP clearly outperform the \ac{SPA} and the CCCP, especially for high $E_\text{b}/N_0$. Consistently with our findings on the Ising graphs, the cycBP algorithm performs better than the extrinsic variant $\text{cycBP}_\text{e}$. The latter can also be trained towards $\mathcal{L}_\text{Bethe}$ without degrading the detection performance. This is particularly surprising since it thereby clearly outperforms the CCCP. The optimization of the cycBP algorithm towards $\mathcal{L}_\text{Bethe}$ does not converge and is therefore not shown in Fig.~\ref{fig:BER_varChannel_N4}. However, since training towards the \ac{BMI} is feasible for large $N$, the supervised training of the cycBP algorithm yields an attractive algorithm with low complexity and superior performance, which can be highly relevant for practical applications.

\subsection{Discussion}
To investigate the two central questions which we formulated in Sec.~\ref{sec:nmp} and to show the potential of our method, we investigated compact models with ${N=4}$. These models are expressive examples for analysis since they have a high density of short cycles and because the true marginals are available as ground truth data.
However, verifying the capability of the proposed cycBP algorithm for practical applications requires extensive numerical evaluation on larger graphs and varying graph structures. This is ongoing work and our preliminary results are promising.

\section{Conclusion}
This work considered message passing for approximate inference and showed the existence of message update rules which perform especially well on cyclic graphs where the \ac{SPA} fails. We challenged the extrinsic information principle for cyclic graphs and proposed an alternative message update rule which also takes intrinsic information into account. The gain was demonstrated by numerical experiments on two exemplary classes of factor graphs. The learned message update rules generalize well and training is extremely fast since the update rule is defined by a very compact \ac{NN} that is reused at all factor nodes. We furthermore proposed a novel unsupervised and application-agnostic loss function that follows the idea of the Bethe approximation.

\begin{acknowledgements} 
This work has received funding in part from the European Research Council (ERC) under the European Union’s Horizon 2020 research and innovation programme (grant agreement No. 101001899) and in part from the German Federal Ministry of Education and Research (BMBF) within the project Open6GHub (grant agreement 16KISK010).
\end{acknowledgements}

\end{document}